\documentclass[journal]{IEEEtran}
\usepackage{amsmath,amsfonts}
\usepackage{algorithmic}
\usepackage{algorithm}
\usepackage{array}
\usepackage[caption=false,font=footnotesize,labelfont=footnotesize,textfont=footnotesize]{subfig}
\usepackage{textcomp}
\usepackage{stfloats}
\usepackage{url}
\usepackage{verbatim}
\usepackage{graphicx}
\usepackage{cite}
\usepackage[table,xcdraw]{xcolor}
\usepackage{booktabs}
\usepackage{multirow}
\usepackage{amssymb}
\usepackage{newtxtext}  
\usepackage{newtxmath}  

\begin{document}

\title{3DResT: A Strong Baseline for Semi-Supervised \\ 3D Referring Expression Segmentation}

\author{Wenxin Chen, Mengxue Qu, Weitai Kang, Yan Yan,  \IEEEmembership{Member, IEEE}, Yao Zhao, \IEEEmembership{Fellow, IEEE} and Yunchao Wei, \IEEEmembership{Member, IEEE}
\thanks{Wenxin Chen, Mengxue Qu, Yao Zhao and Yunchao Wei are with BeijingKey Laboratory of Advanced Information Science and Network and Instituteof Information Science, Beijing Jiaotong University, Beijing, 100044, China (email: chenwenxin@bitu.edu.cn; qumengxue@bitu.edu.cn; yzhao@bitu.edu.cn; yunchao.wei@bitu.edu.cn).}
\thanks{Weitai Kang and Yan Yan are with University of Illinois Chicago, Chicago, Illinois 60607 USA (email: wkang126@uic.edu; yyan55@uic.edu).}
}



\maketitle

\begin{abstract}
3D Referring Expression Segmentation (3D-RES) typically requires extensive instance-level annotations, which are time-consuming and costly. 
Semi-supervised learning (SSL) mitigates this by using limited labeled data alongside abundant unlabeled data, improving performance while reducing annotation costs.
SSL uses a teacher–student paradigm where teacher generates high-confidence–filtered pseudo-labels to guide student. However, in the context of 3D-RES, where each label corresponds to a single mask and labeled data is scarce, existing SSL methods treat high-quality pseudo-labels merely as auxiliary supervision, which limits the model's learning potential. The reliance on high-confidence thresholds for filtering often results in potentially valuable pseudo-labels being discarded, restricting the model's ability to leverage the abundant unlabeled data.
Therefore, we identify two critical challenges in semi-supervised 3D-RES, namely, inefficient utilization of high-quality pseudo-labels and wastage of useful information from low-quality pseudo-labels. In this paper, we introduce the first semi-supervised learning framework for 3D-RES, presenting a robust baseline method named \textbf{3DResT}. To address these challenges, we propose two novel designs called Teacher-Student Consistency-Based Sampling (TSCS) and Quality-Driven Dynamic Weighting (QDW). TSCS aids in the selection of high-quality pseudo-labels, integrating them into the labeled dataset to strengthen the labeled supervision signals. 
QDW preserves low-quality pseudo-labels by dynamically assigning them lower weights, allowing for the effective extraction of useful information rather than discarding them.
Extensive experiments conducted on the widely used benchmark 
demonstrate the effectiveness of our method. Notably, with only 1\% labeled data, 3DResT achieves an mIoU improvement of 8.34 points compared to the fully supervised method.
\end{abstract}

\begin{IEEEkeywords}
Referring Expression Segmentation, 3D Point Cloud, Semi-Supervised Learning.
\end{IEEEkeywords}

\section{Introduction}
\IEEEPARstart{3}D Referring Expression Segmentation (3D-RES) aims to segment objects in 3D scenes based on natural language descriptions.
It provides 
a wide range of applications in robotics \cite{ref53, ref54}, virtual reality \cite{ref55, ref56}, human-computer interaction \cite{ref57, ref58}, and other domains. 
Compared with conventional 3D segmentation tasks \cite{ref1, ref2, ref3, ref4, ref5, ref6, ref7, ref8, ref49} , 3D-RES \cite{ref9, ref10, ref11, ref12, ref13, ref14, ref15} requires not only the precise geometric segmentation of 3D point cloud data but also the comprehension of the target object’s location and attributes as described in natural language. These requirements increase the complexity of semantic comprehension. As a result, 3D-RES faces a higher demand for annotated data, as the diversity and complexity of language descriptions make acquiring such data more challenging than in traditional 3D segmentation tasks.

\begin{figure}[!t]
\centering
\includegraphics[width=3.5in]{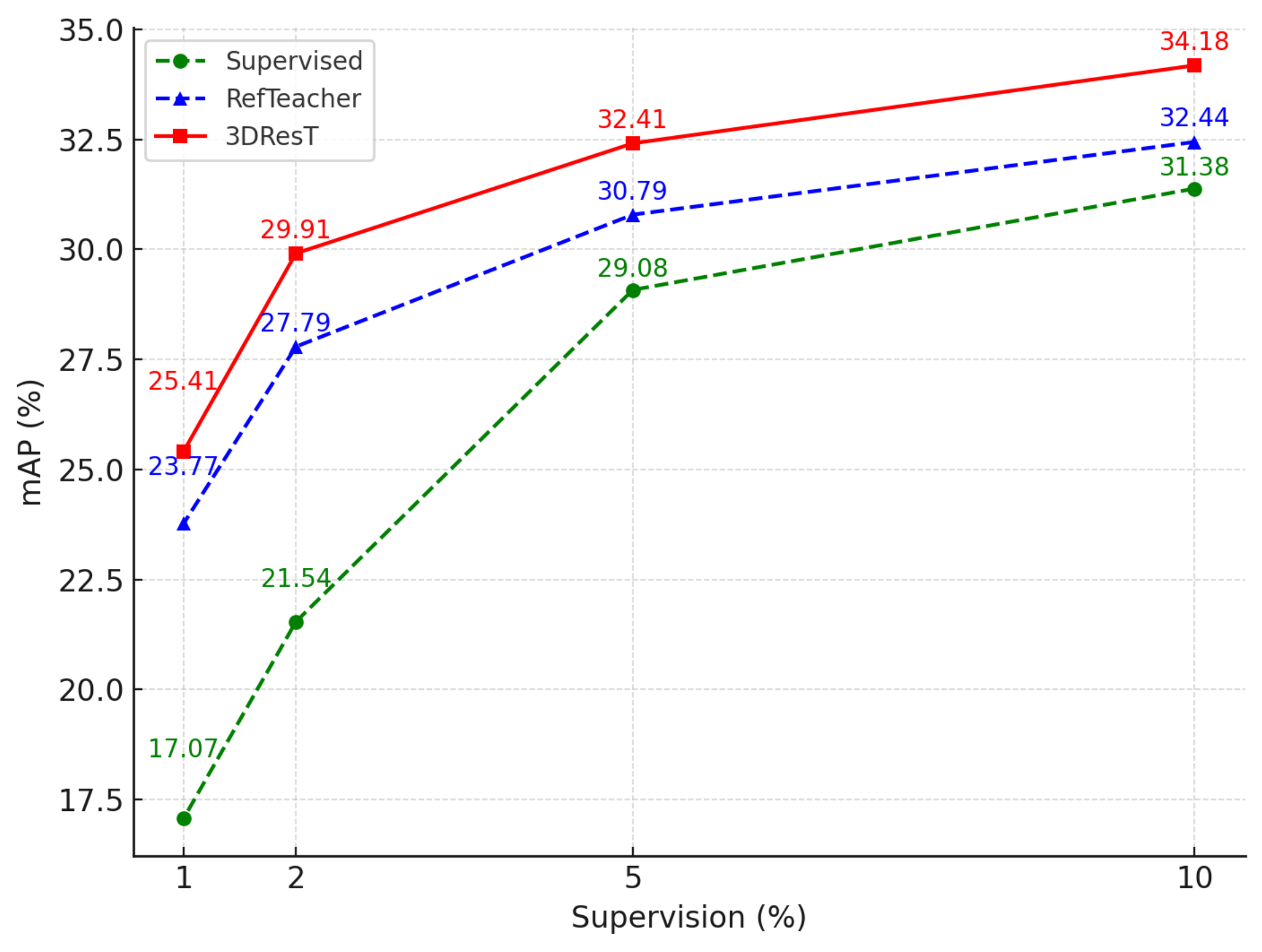}
\caption{Our proposed model can efficiently leverage the unlabeled data and perform favorably against the fully supervised method and the existing semi-supervised method in 2D referring work. Supervised method only uses labeled supervision, while RefTeacher and 3DResT leverage both labeled supervision and unlabeled data.}
\label{fig1}
\end{figure}

3D point cloud data \cite{ref16, ref17} is inherently sparse and high-dimensional, which makes instance-level annotation both time-consuming and labor-intensive, with costs significantly exceeding those of 2D image annotations \cite{ref18, ref19, ref20}. In practical applications, the collection of annotated data is often constrained, and many scenarios lack large-scale annotated datasets, making it crucial to effectively leverage the limited annotations. Moreover, in the 3D-RES task, each label provides only one instance-level supervision \cite{ref21}, compounding the scarcity of supervision and restricting the performance of existing methods. 
The majority of current 3D-RES methods \cite{ref9, ref10, ref11, ref12, ref13, ref14, ref15} are based on fully supervised learning, which requires a large number of annotated data during training. However, these approaches heavily rely on the annotations, making them unsuitable for real-world scenarios where annotation resources are limited. As a result, semi-supervised learning (SSL) \cite{ref22, ref23, ref24, ref25, ref26, ref27, ref28, ref29, ref30, ref31, ref32, ref33, ref59} 
has emerged as an effective solution, achieving significant progress in various computer vision tasks in recent years. Especially in the data-scarce 3D domain, SSL has shown promising results in tasks such as 3D object detection \cite{ref27, ref28} and 3D segmentation \cite{ref26, ref29, ref30, ref31, ref32, ref33}. By combining a small amount of labeled data with a large quantity of unlabeled data, SSL can effectively reduce the dependence on large-scale annotated datasets. A typical SSL method employs a Teacher-Student framework \cite{ref22, ref27, ref28, ref30, ref31, ref32, ref33}, where the teacher network generates pseudo-labels and the student network then utilizes these pseudo-labels to train on unlabeled data. This approach effectively harnesses the unlabeled data to enhance model performance in scenarios with limited annotated data.
However, directly transferring such SSL methods to 3D-RES still suffers from two major issues: (1) Inefficient utilization of high-quality pseudo-labels: Current SSL methods \cite{ref22, ref27, ref28, ref30, ref31, ref32, ref33} primarily use pseudo-labels in the loss computation for unlabeled data. Although high-quality pseudo-labels can be beneficial for model optimization, they are typically employed merely as auxiliary supervision and fail to effectively augment the labeled dataset, thereby constraining the model's learning capacity. This issue is further compounded in 3D-RES, where each label provides only one instance-level supervision, amplifying the scarcity of supervision.
(2) Wastage of useful information from low-quality pseudo-labels: Traditional SSL methods often rely on high-confidence thresholds to filter pseudo-labels. This suboptimal strategy results in many pseudo-labels being discarded in semi-supervised 3D-RES. Especially when the overall quality of pseudo-labels is low, this filtering method further restricts the model's learning potential. As a result, utilizing the useful information from low-quality pseudo-labels becomes crucial.

Based on these observations, we propose the first semi-supervised learning framework for the 3D-RES task, called \textbf{3DResT}. 
To solve the inefficient utilization of high-quality pseudo-labels, we introduce a Teacher-Student Consistency-Based Sampling (TSCS) based on Teacher-Student consistency. 
We assess the reliability of pseudo-labels through Teacher-Student consistency and integrate them into the labeled dataset only when the predictions from both networks are highly consistent, such as when the IoU exceeds a threshold of 0.95. Therefore, high-quality pseudo-labels are leveraged to expand the labeled dataset, addressing the scarcity of supervision and optimizing the utilization of them.
Additionally, to avoid discarding low-quality pseudo-labels 
due to an overly strict high-confidence threshold, we propose a Quality-Driven Dynamic Weighting (QDW) that assigns weights dynamically based on the quality of pseudo-labels. Even though the low-quality pseudo-labels can still contribute to model training with lower weights, thereby reducing the wastage of useful information. By leveraging both strategies, 3DResT fully exploits high-quality pseudo-labels and low-quality pseudo-labels, resulting in significant improvements in the 3D-RES task.

To validate 3DResT, we conduct extensive quantitative and qualitative experiments on ScanRefer \cite{ref21}. The experimental results demonstrate that 3DResT significantly outperforms the supervised baselines, achieving an 8.34\% improvement on 1\% of ScanRefer, as shown in Fig. \ref{fig1}.

Overall, the contributions of this paper are threefold:
\begin{itemize}
    \item{We propose the first attempt of semi-supervised learning for 3D-RES with a strong baseline method called 3DResT.}
    \item{We identify two challenges of semi-supervised 3D-RES, \textit{i.e}. inefficient utilization of high-quality pseudo-labels and wastage of useful information from low-quality pseudo-labels, and address them with two novel designs, namely Teacher-Student Consistency-Based Sampling (TSCS) and Quality-Driven Dynamic Weighting (QDW).}
    \item{3DResT achieves significant performance improvements on ScanRefer dataset over the fully supervised methods.}
\end{itemize}

\section{Related Work}
\subsection{3D Referring Expression Comprehension and Segmentation}
3D Referring Expression Comprehension (3D-REC) \cite{ref37, ref38, ref39, ref40, ref41, ref42} has become an essential task in 3D scene understanding, aiming to localize target objects in 3D environments based on natural language descriptions. The benchmarks like ScanRefer \cite{ref21}, Nr3D \cite{ref17}, Sr3D \cite{ref17} have significantly advanced this field by introducing datasets that feature detailed object categories and complex scenes with multiple instances.

Building on 3D-REC, 3D Referring Expression Segmentation (3D-RES) further enhances the task by not only localizing objects but also generating precise 3D masks to accurately represent the referred objects. Early approaches, such as TGNN \cite{ref15}, employed a two-stage framework that relied on instance proposals. 3D-STMN \cite{ref14} introduced a one-stage approach, significantly improving the inference speed. To enhance the performance of 3D-RES, significant works \cite{ref9, ref10, ref11, ref12, ref13, ref43} have focused on exploring the fusion of visual and linguistic information. However, during training, both 3D-REC and 3D-RES heavily require a large amount of annotated data, which is time-consuming and costly. To tackle these challenges, we utilize semi-supervised learning to alleviate the dependence on large-scale annotated datasets.

\subsection{Semi-Supervised Learning}
Semi-Supervised Learning (SSL) has achieved significant progress in various computer vision tasks \cite{ref22, ref23, ref24, ref25, ref26, ref27, ref28, ref29, ref30, ref31, ref32, ref33, ref59}. SSL methods are broadly categorized into two main approaches: consistency-based methods \cite{ref23, ref24, ref25, ref26, ref29}, which leverage regularization losses to ensure that predictions from the teacher and student networks remain consistent under varying noisy inputs, and pseudo-label-based methods \cite{ref22, ref27, ref28, ref30, ref31, ref32, ref33}, which employ a well-trained teacher network to generate pseudo-labels for unlabeled data to optimize the student network.

Building on these advancements, there are massive studies \cite{ref26, ref29, ref30, ref31, ref32, ref33} on semi-supervised 3D segmentation. Besides, recent studies \cite{ref44, ref45, ref46} have extended SSL to referring expression comprehension and segmentation. A notable example is RefTeacher \cite{ref46}, using the pseudo-label-based methods, presents the first attempt of semi-supervised learning for the 2D referring task. However, due to the increasing difficulty of acquiring 3D data and the higher annotation costs compared to 2D data, the scale of 3D datasets is significantly smaller than that of 2D datasets. In this paper, we apply the Teacher-Student Consistency-Based Sampling (TSCS) that facilitates the selection of high-quality pseudo-labels and incorporates them into the labeled dataset to expand the labeled supervision signals, which maximizes the utilization of high-quality pseudo-labels.

\subsection{Active Learning}
Several active learning approaches have been developed for label sampling \cite{ref28, ref34, ref35, ref36, ref59}. For example, CALD \cite{ref35} measures information through data consistency, while MI-AOD \cite{ref36} leverages multi-instance learning to reduce noise in pseudo-labels. Additionally, Active-Teacher \cite{ref28} incorporates a teacher-student framework to enhance the data sampling process. In this paper, we apply active learning merely based on the predictions of the teacher and student networks, without incorporating any additional predictions.
\begin{figure*}[!t]
\centering
\includegraphics[width=7.4in]{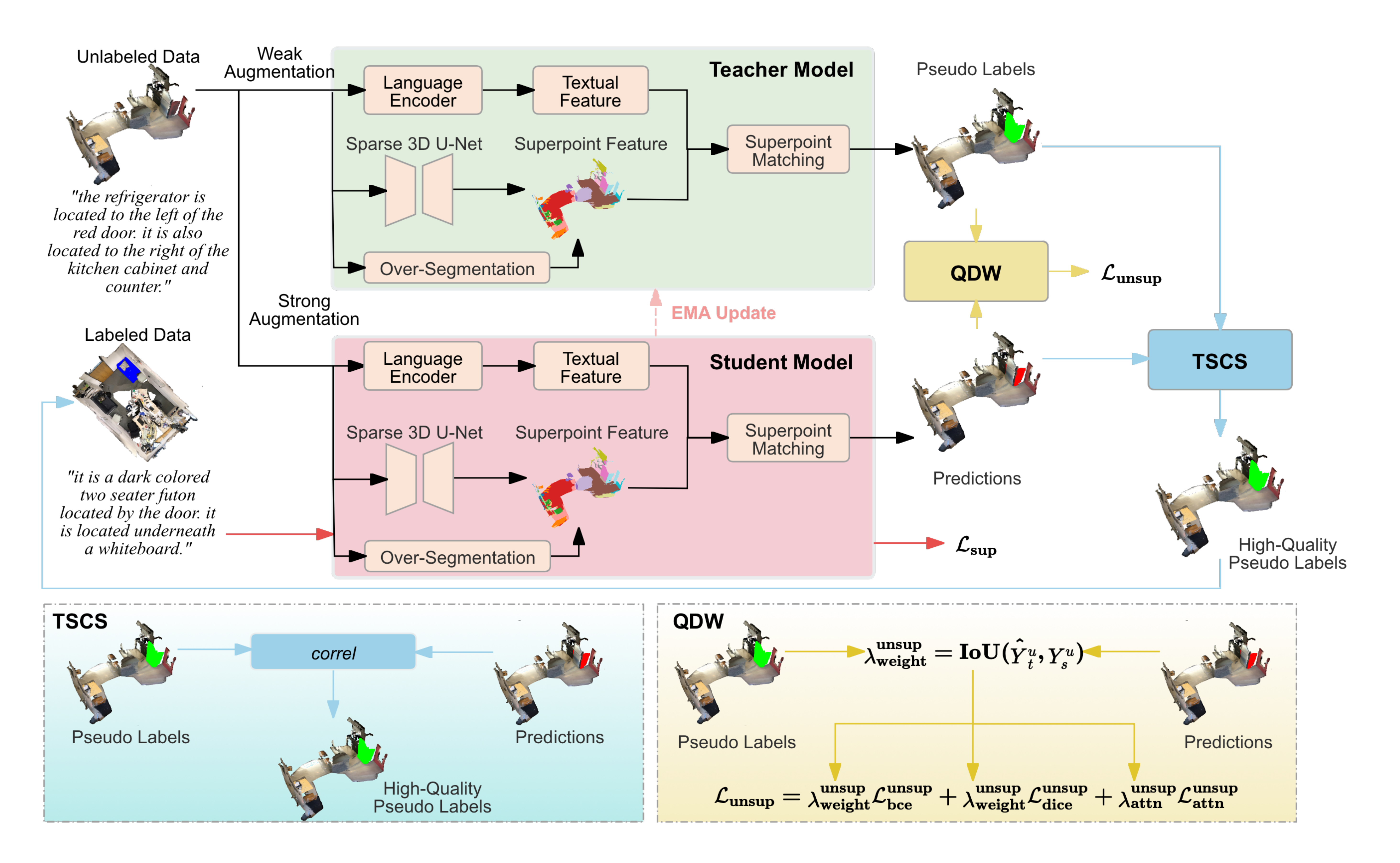}
\caption{The overall semi-supervised 3D-RES framework, 3DResT, consists of two 3D-RES networks with identical configurations, referred to as the Teacher and Student. The Teacher predicts pseudo-labels for unlabeled data, which are used to train the Student alongside a small number of labeled samples. The Teacher is updated via EMA \cite{ref46} from the Student. Additionally, Teacher-Student Consistency-Based Sampling (TSCS) and Quality-Driven Dynamic Weighting (QDW) are employed to address the challenges of inefficient utilization of high-quality pseudo-labels and wastage of useful information from low-quality pseudo-labels. The \textit{correl} in TSCS is a metric helping to better select high-quality pseudo-labels which will be described in Section \ref{III-C} later.}
\label{fig2}
\end{figure*}

\section{Methodology}
In this section, we provide a comprehensive overview of 3DResT. First, we present a task definition of our work in Section \ref{III-A}. Next, we introduce the overall framework of 3DResT in Section \ref{III-B}. Finally, we detail the proposed Teacher-Student Consistency-Based Sampling module and Quality-Driven Dynamic Weighting module in Section \ref{III-C} and Section \ref{III-D}, respectively.
\subsection{Task Definition}
\label{III-A}
Given a set of labeled data $D_l = \{(V_i^l, T_i^l), Y_i^l\}_{i=1}^{N_l}$ and unlabeled data $D_u = \{(V_i^u, T_i^u)\}_{i=1}^{N_u}$, the objective of semi-supervised 3D-RES can be formulated as
\begin{equation} 
    \label{equation:1}
    \min \mathcal{L}(\theta; D_l, D_u),
\end{equation}
where $\mathcal{L}$ represents the semi-supervised objective function. $\theta$ denotes the 3D-RES model. $V$, $T$ and $Y$ correspond to the 3D point clouds, textual descriptions and mask annotations, respectively. In practice, the number of labeled data is significantly less than the unlabeled data, satisfying $N_l \ll N_u$.

As 3D-RES is a language-driven visual segmentation task, the absence of textual information will render the model unable to generate predictions. Besides, such language-level annotations are relatively easy to acquire from existing vision-language datasets \cite{ref47} and online resources. In this case, in our semi-supervised framework, we discard the mask annotations while retaining the textual descriptions.
\subsection{3DResT's Framework}
\label{III-B}
We first present the overall framework of 3DResT designed for semi-supervised 3D-RES. As illustrated in Fig. \ref{fig2}, our framework comprises two 3D-RES networks with identical network configurations, serving as the Teacher and the Student, respectively. During training, the teacher network creates pseudo-labels for the unlabeled data, and the student network is trained using both the ground-truth labels and the pseudo-labels. In this case, the overall optimization objective for the student is defined by
\begin{equation}
    \label{equation:2}
    \mathcal{L} = \mathcal{L}_{\text{sup}} + \lambda_u \mathcal{L}_{\text{unsup}},
\end{equation}
where $\lambda_u$ is a hyper-parameter that controls the weight of the unsupervised loss. $\mathcal{L}_{\text{sup}}$ is the loss function of the 3D-RES model \cite{ref14}, which can be defined by
\begin{equation}
    \label{equation:3}
    \mathcal{L}_{\text{sup}} = \lambda_{\text{bce}}^{\text{sup}}\mathcal{L}_{\text{bce}}^{\text{sup}} + \lambda_{\text{dice}}^{\text{sup}}\mathcal{L}_{\text{dice}}^{\text{sup}} + \lambda_{\text{rel}}^{\text{sup}}\mathcal{L}_{\text{rel}}^{\text{sup}} + \lambda_{\text{score}}^{\text{sup}}\mathcal{L}_{\text{score}}^{\text{sup}}.
\end{equation}

To enhance the effectiveness of semi-supervised 3D-RES, we incorporate several novel designs of SSL \cite{ref31, ref33, ref44, ref45, ref46} into 3DResT. These include the burn-in stage \cite{ref46} and exponential moving average (EMA) \cite{ref31, ref33, ref46}. Specifically, before initiating the teacher-student mutual learning process, the teacher network is first trained on labeled data for a short period, which is called burn-in stage \cite{ref46}. This stage makes the teacher network equipped with preliminary segmentation capabilities, enabling it to generate reliable pseudo-labels. Following the burn-in stage, the parameters of the teacher network are transferred to initialize the student network.

During the teacher-student mutual learning stage, the student network is trained using both labeled and unlabeled data, as defined in (\ref{equation:2}). In contrast, the teacher network does not perform gradient backpropagation. Instead, the teacher network parameters are updated from the student via EMA \cite{ref31, ref33, ref46}.
The optimization objectives for the teacher and student networks are formulated as
\begin{align}
    \label{equation:4}
    \theta_s^i &\leftarrow \theta_s^i + \gamma \frac{\partial \big( \mathcal{L}_{\text{sup}} + \lambda_u \mathcal{L}_{\text{unsup}} \big)}{\partial \theta_s^i},  \\
    \label{equation:5}
    \theta_t^i &\leftarrow \alpha \theta_t^{i-1} + (1 - \alpha) \theta_s^i, 
\end{align}
where $\theta_t$ and $\theta_s$ represent the parameters of the teacher and student networks, respectively. $i$ denotes the training iteration, and $\alpha$ is the keeping rate. This update mechanism effectively prevents the teacher from overfitting to the limited labeled data \cite{ref51}. Overall, the complete workflow of 3DResT is detailed in Algorithm \ref{algorithm:1}.

\begin{algorithm}[!t]
    \caption{Pseudo Code of 3DResT}
    \label{algorithm:1}
    \textbf{Input}: Labeled Data $D_l = \{(V_i^l, T_i^l), Y_i^l\}_{i=1}^{N_l}$ and Unlabeled Data $D_u = \{(V_i^u, T_i^u)\}_{i=1}^{N_u}$, burn-in iteration $N_1$, maximum iteration $N_2$, epoch of sampling $K$, filter threshold $s$ \newline
    \textbf{Output}: Teacher Model $\theta_t^i$ and Student Model $\theta_s^i$
    \begin{algorithmic} 
        \FOR{$i < N_2$}
            \IF{$i < N_1$}
                \STATE Update $\theta_t^i$ on $D_l$ by (\ref{equation:3})
            \ENDIF
            \IF{$i = N_1$}
                \STATE Initialize $\theta_s^i$ with $\theta_t^i$
            \ENDIF
            \IF{$i > N_1$}
                \STATE Predict pseudo-labels $\{(V_i^u, T_i^u), \hat{Y}_i^u\}_{i=1}^{N_u}$ by Teacher $\theta_t^i$
                \STATE Calculate $\mathcal{L}_{\text{sup}}$ on $D_l$ by (\ref{equation:3})
                \STATE Calculate $\mathcal{L}_{\text{unsup}}$ on $D_u$ by (\ref{equation:11})
                \STATE Update $\theta_s^i$ and $\theta_t^i$ by (\ref{equation:4}) and (\ref{equation:5})
            \ENDIF
            \IF{$epoch = K$}
                \STATE Calculate the $correl$ on $\{(V_i^u, T_i^u), \hat{Y}_i^u\}_{i=1}^{N_u}$ by (\ref{equation:6})
                \STATE Select pseudo label  $\{(V_i^p, T_i^p), \hat{Y}_i^p\}_{i=1}^{N_p}$ with $s$ and $correl$
                \STATE Update labeled set $D_l = D_l \cup \{(V_i^p, T_i^p), \hat{Y}_i^p\}_{i=1}^{N_p}$
                \STATE Update unlabeled set $D_u = D_u - \{(V_i^p, T_i^p)\}_{i=1}^{N_p}$
            \ENDIF
        \ENDFOR
        \RETURN $\theta_t^i$, $\theta_s^i$
    \end{algorithmic}
\end{algorithm}

As shown in (\ref{equation:2}), the unsupervised loss serves as an auxiliary supervision signal for the supervised loss to fully harness the model’s potential. Existing semi-supervised segmentation methods \cite{ref31, ref32, ref33} typically leverage pseudo-labels merely as auxiliary supervision and rely on solid high-confidence thresholds to filter them, which discard the potentially valuable pseudo-labels. These cases lead to two key issues for semi-supervised 3D-RES. The first is the inefficient utilization of high-quality pseudo-labels. The second is the wastage of useful information from low-quality pseudo-labels.

\subsection{Teacher-Student Consistency-Based Sampling}
\label{III-C}
To address the issue of inefficient utilization of high-quality pseudo-labels, we first propose the Teacher-Student Consistency-Based Sampling (TSCS). Its main principle is to select high-quality pseudo-labels and add them to the labeled dataset to continuously influence the model’s subsequent training.

Given unlabeled data, the common approach in existing methods \cite{ref22, ref27, ref28, ref30, ref31, ref32, ref33} is that the teacher network generates pseudo-labels while the student network produces corresponding predictions, followed by calculating the unsupervised loss to optimize the student network. However, the pseudo-labels are only treated as auxiliary supervision.

To effectively select high-quality pseudo-labels, we calculate the correlation score between the teacher’s and student’s mask predictions. Pseudo-labels with correlation scores exceeding a specified high threshold are treated as ground truth for the corresponding unlabeled samples and subsequently incorporated into the labeled dataset, which is outlined in Algorithm \ref{algorithm:1}. The correlation score helps to better select high-quality pseudo-labels, expand the size of the labeled set and enable the model efficiently utilize the high-quality pseudo-labels, which is defined as
\begin{equation}
    \label{equation:6}
    correl = \frac{\left| Y_s^u \cap \hat{Y}_t^u\right|}{\left|Y_s^u \cup \hat{Y}_t^u\right|},
\end{equation}
where $Y_s^u$ represents the mask prediction of the student model and $\hat{Y}_t^u$ denotes the pseudo-label mask prediction of the teacher model. As long as the correlation score exceeds a threshold, the pseudo-label is considered as the high-quality pseudo-label.

\subsection{Quality-Driven Dynamic Weighting}
\label{III-D}
Since we cannot directly apply traditional filtering strategies, which significantly compromise the already limited pseudo-labels and thus waste the useful information of low-quality pseudo-labels. To tackle this issue, we introduce Quality-Driven Dynamic Weighting (QDW), which utilizes $\lambda_{\text{weight}}^{\text{unsup}}$ to dynamically calculate the IoU between the teacher’s output and the student’s output to adjust the weight of the unsupervised loss for each sample. To be specific, QDW can be defined as
\begin{equation}
    \label{equation:7}
    \lambda_{\text{weight}}^{\text{unsup}} = \text{IoU}(Y_s^u, \hat{Y}_t^u),
\end{equation}
where \text{IoU} represents the intersection over union. $Y_s^u$ and $\hat{Y}_t^u$ represent the mask predictions of the student model and the teacher model, respectively. When the teacher predicts a low-quality pseudo-label, QDW can dynamically adjust its weight instead of discarding it. This approach enables the optimal extraction of potentially useful information from low-quality pseudo-labels.

For the learning of unlabeled data, there are some details of unsupervised loss.
We use the $\mathcal{L}_{\text{bce}}^{\text{unsup}}$ for the mask predictions:
\begin{equation}
    \label{equation:8}
    \mathcal{L}_{\text{bce}}^{\text{unsup}} = \text{BCE}(Y_s^u, \hat{Y}_t^u).
\end{equation}

Following $\mathcal{L}_{\text{sup}}$ \cite{ref48}, we also use the $\mathcal{L}_{\text{dice}}^{\text{unsup}}$ to balance the foreground-background:
\begin{equation}
    \label{equation:9}
    \mathcal{L}_{\text{dice}}^{\text{unsup}} = 1 - \frac{2 \sum_{i=1}^{N_u} Y_s^i \hat{Y}_t^i}{\sum_{i=1}^{N_u} Y_s^i + \sum_{i=1}^{N_u} \hat{Y}_t^i}.
\end{equation}

For fair comparison to RefTeacher \cite{ref46}, we also use the $\mathcal{L}_{\text{attn}}^{\text{unsup}}$ and the corresponding hyperparameter $\lambda_{\text{attn}}^{\text{unsup}}$ to facilitate the vision-language alignment:
\begin{equation}
    \label{equation:10}
    \mathcal{L}_{\text{attn}}^{\text{unsup}} = \sum_{i=1}^{n} \frac{1}{n} \left( \hat{F}_{\text{att}}^i - F_{\text{att}}^i \right)^2,
\end{equation}
where $n = \mathcal{N}_s \times d$. $\mathcal{N}_s$ denotes the number of the superpoints \cite{ref7}. $\hat{F}_{\text{att}}$ and $F_{\text{att}}$ denote the attention features of the teacher and student, respectively.

Together, the unsupervised loss can be formulated as
\begin{equation}
    \label{equation:11}
    \mathcal{L}_{\text{unsup}} = \lambda_{\text{weight}}^{\text{unsup}}\mathcal{L}_{\text{bce}}^{\text{unsup}} + \lambda_{\text{weight}}^{\text{unsup}}\mathcal{L}_{\text{dice}}^{\text{unsup}} + \lambda_{\text{attn}}^{\text{unsup}}\mathcal{L}_{\text{attn}}^{\text{unsup}},
\end{equation}
where $\lambda_{\text{weight}}^{\text{unsup}}$ can dynamically adjust the weight of the unsupervised loss for each sample, $\lambda_{\text{attn}}^{\text{unsup}}$ are hyperparameters used to balance the $\mathcal{L}_{\text{attn}}^{\text{unsup}}$.

\section{Experiments}
\subsection{Experiment Settings}
\label{IV-A}
All models are trained using AdamW \cite{ref52} with a constant learning rate of $0.0001$. The batch size is set to 64 labeled and 64 unlabeled vision-text pairs. The total training steps are 13k, with burn-in steps set to 2k by default. The hyperparameters $\alpha$, $s$ and  $\lambda_u$
are set to 0.9996, 0.95 and 0.5, respectively.
For strong-weak augmentations, we use \text{RandomResize}, \text{RandomSizeCrop} and \text{ColorJitter} as strong augmentations, while \text{Rotation} is used as the weak augmentation. All experiments are implemented 
on two NVIDIA 4090 GPUs. For other network configurations, such as text encoder and visual backbones, we follow the default settings of 3D-STMN \cite{ref14}.
\begin{table*}
    \centering
    \caption{Comparisons of 3DResT and Baselines on ScanRefer.
    For all approaches, we use 3D-STMN \cite{ref14} as the 3D-RES model.
    Supervised denotes the fully supervised training.
    RefTeacher \cite{ref46} is the semi-supervised approach of 2D referring task.}
    \label{tab1}

    \begin{tabular*}{\linewidth}{@{\extracolsep{\fill}}l c c c c c c c c c@{}}
        \toprule
        \multicolumn{10}{c}{\textbf{1\% Labeled Data}} \\ 
        \midrule
        \multirow{2}{*}{\textbf{Method}} 
        & \multicolumn{3}{c}{\textbf{Unique ($\sim$19\%)}} 
        & \multicolumn{3}{c}{\textbf{Multiple ($\sim$81\%)}} 
        & \multicolumn{3}{c}{\textbf{Overall}} \\ 
        \cmidrule(lr){2-4} \cmidrule(lr){5-7} \cmidrule(lr){8-10}
        & \textbf{0.25} & \textbf{0.5} & \textbf{mIoU}
        & \textbf{0.25} & \textbf{0.5} & \textbf{mIoU}
        & \textbf{0.25} & \textbf{0.5} & \textbf{mIoU} \\
        \midrule
        Supervised 
        & \small 48.67 & \small 40.16 & \small 37.42
        & \small 18.43 & \small 10.54 & \small 12.17
        & \small 24.30 & \small 16.29 & \small 17.07 \\
        
        RefTeacher 
        & \small 65.04\tiny+16.37 & \small 54.74\tiny+14.58 & \small 50.05\tiny+12.63
        & \small 27.48\tiny+9.05  & \small 13.53\tiny+2.99 & \small 17.45\tiny+5.28
        & \small 34.77\tiny+10.47 & \small 21.53\tiny+5.24 & \small 23.77\tiny+6.70 \\
        
        3DResT 
        & \colorbox[HTML]{FFD966}{\small 70.30\tiny+21.63}
        & \colorbox[HTML]{FFD966}{\small 60.92\tiny+20.76}
        & \colorbox[HTML]{FFD966}{\small 55.42\tiny+18.00}
        & \colorbox[HTML]{FFD966}{\small 28.80\tiny+10.37}
        & \colorbox[HTML]{FFD966}{\small 14.30\tiny+3.76}
        & \colorbox[HTML]{FFD966}{\small 18.18\tiny+6.01}
        & \colorbox[HTML]{FFD966}{\small 36.85\tiny+12.55}
        & \colorbox[HTML]{FFD966}{\small 23.35\tiny+7.06}
        & \colorbox[HTML]{FFD966}{\small 25.41\tiny+8.34} \\
        \bottomrule
    \end{tabular*}
    \vspace{0.02cm}

    \begin{tabular*}{\linewidth}{@{\extracolsep{\fill}}l c c c c c c c c c@{}}
        \toprule
        \multicolumn{10}{c}{\textbf{2\% Labeled Data}} \\ 
        \midrule
        \multirow{2}{*}{\textbf{Method}} 
        & \multicolumn{3}{c}{\textbf{Unique ($\sim$19\%)}} 
        & \multicolumn{3}{c}{\textbf{Multiple ($\sim$81\%)}} 
        & \multicolumn{3}{c}{\textbf{Overall}} \\ 
        \cmidrule(lr){2-4} \cmidrule(lr){5-7} \cmidrule(lr){8-10}
        & \textbf{0.25} & \textbf{0.5} & \textbf{mIoU}
        & \textbf{0.25} & \textbf{0.5} & \textbf{mIoU}
        & \textbf{0.25} & \textbf{0.5} & \textbf{mIoU} \\
        \midrule
        Supervised 
        & \small 60.22 & \small 50.30 & \small 46.50
        & \small 23.99 & \small 12.76 & \small 15.53
        & \small 31.02 & \small 20.05 & \small 21.54 \\
        
        RefTeacher 
        & \small 75.56\tiny+15.34 & \small 66.99\tiny+16.69 & \small 60.27\tiny+13.77
        & \small 31.24\tiny+7.25  & \small 15.80\tiny+3.04  & \small 19.97\tiny+4.44
        & \small 39.84\tiny+8.82  & \small 25.74\tiny+5.69 & \small 27.79\tiny+6.25 \\
        
        3DResT 
        & \colorbox[HTML]{FFD966}{\small 80.54\tiny+20.32}
        & \colorbox[HTML]{FFD966}{\small 74.53\tiny+24.23}
        & \colorbox[HTML]{FFD966}{\small 66.01\tiny+19.51}
        & \colorbox[HTML]{FFD966}{\small 32.27\tiny+8.28}
        & \colorbox[HTML]{FFD966}{\small 18.88\tiny+6.12}
        & \colorbox[HTML]{FFD966}{\small 21.22\tiny+5.69}
        & \colorbox[HTML]{FFD966}{\small 41.61\tiny+10.59}
        & \colorbox[HTML]{FFD966}{\small 29.68\tiny+9.63}
        & \colorbox[HTML]{FFD966}{\small 29.91\tiny+8.37} \\
        \bottomrule
    \end{tabular*}
    \vspace{0.02cm}

    \begin{tabular*}{\linewidth}{@{\extracolsep{\fill}}l c c c c c c c c c@{}}
        \toprule
        \multicolumn{10}{c}{\textbf{5\% Labeled Data}} \\ 
        \midrule
        \multirow{2}{*}{\textbf{Method}} 
        & \multicolumn{3}{c}{\textbf{Unique ($\sim$19\%)}} 
        & \multicolumn{3}{c}{\textbf{Multiple ($\sim$81\%)}} 
        & \multicolumn{3}{c}{\textbf{Overall}} \\ 
        \cmidrule(lr){2-4} \cmidrule(lr){5-7} \cmidrule(lr){8-10}
        & \textbf{0.25} & \textbf{0.5} & \textbf{mIoU}
        & \textbf{0.25} & \textbf{0.5} & \textbf{mIoU}
        & \textbf{0.25} & \textbf{0.5} & \textbf{mIoU} \\
        \midrule
        Supervised 
        & \small 79.51 & \small 70.41 & \small 63.30
        & \small 33.52 & \small 16.87 & \small 20.84
        & \small 42.45 & \small 27.26 & \small 29.08 \\
        
        RefTeacher 
        & \small 78.70\tiny-0.81  & \small 71.82\tiny+1.41 & \small 64.22\tiny+0.92
        & \small 33.93\tiny+0.41 
        & \colorbox[HTML]{FFD966}{\small 20.12\tiny+3.25}
        & \small 22.74\tiny+1.90
        & \small 42.62\tiny+0.17 & \small 30.15\tiny+2.89 & \small 30.79\tiny+1.71 \\
        
        3DResT 
        & \colorbox[HTML]{FFD966}{\small 82.82\tiny+3.31}
        & \colorbox[HTML]{FFD966}{\small 76.04\tiny+5.63}
        & \colorbox[HTML]{FFD966}{\small 67.77\tiny+4.47}
        & \colorbox[HTML]{FFD966}{\small 37.34\tiny+3.82}
        & \small 19.69\tiny+2.82 
        & \colorbox[HTML]{FFD966}{\small 23.90\tiny+3.06}
        & \colorbox[HTML]{FFD966}{\small 46.16\tiny+3.71}
        & \colorbox[HTML]{FFD966}{\small 30.63\tiny+3.37}
        & \colorbox[HTML]{FFD966}{\small 32.41\tiny+3.33} \\
        \bottomrule
    \end{tabular*}
    \vspace{0.02cm}

    \begin{tabular*}{\linewidth}{@{\extracolsep{\fill}}l c c c c c c c c c@{}}
        \toprule
        \multicolumn{10}{c}{\textbf{10\% Labeled Data}} \\ 
        \midrule
        \multirow{2}{*}{\textbf{Method}} 
        & \multicolumn{3}{c}{\textbf{Unique ($\sim$19\%)}} 
        & \multicolumn{3}{c}{\textbf{Multiple ($\sim$81\%)}} 
        & \multicolumn{3}{c}{\textbf{Overall}} \\ 
        \cmidrule(lr){2-4} \cmidrule(lr){5-7} \cmidrule(lr){8-10}
        & \textbf{0.25} & \textbf{0.5} & \textbf{mIoU}
        & \textbf{0.25} & \textbf{0.5} & \textbf{mIoU}
        & \textbf{0.25} & \textbf{0.5} & \textbf{mIoU} \\
        \midrule
        Supervised 
        & \small 84.12 & \small 76.31 & \small 68.27
        & \small 35.80 & \small 18.07 & \small 22.50
        & \small 45.17 & \small 29.38 & \small 31.38 \\
        
        RefTeacher 
        & \small 81.84\tiny-2.28 & \small 77.29\tiny+0.98 & \small 67.51\tiny-0.76
        & \small 34.82\tiny-0.98 
        & \colorbox[HTML]{FFD966}{\small 22.22\tiny+4.15}
        & \small 24.00\tiny+1.50
        & \small 43.94\tiny-1.23 
        & \colorbox[HTML]{FFD966}{\small 32.91\tiny+3.53}
        & \small 32.44\tiny+1.06 \\
        
        3DResT 
        & \colorbox[HTML]{FFD966}{\small 84.93\tiny+0.81}
        & \colorbox[HTML]{FFD966}{\small 79.46\tiny+3.15}
        & \colorbox[HTML]{FFD966}{\small 70.27\tiny+2.00}
        & \colorbox[HTML]{FFD966}{\small 39.20\tiny+3.40}
        & \small 21.62\tiny+3.55 
        & \colorbox[HTML]{FFD966}{\small 25.49\tiny+2.99}
        & \colorbox[HTML]{FFD966}{\small 48.08\tiny+2.91}
        & \small 32.85\tiny+3.47
        & \colorbox[HTML]{FFD966}{\small 34.18\tiny+2.80} \\
        \bottomrule
    \end{tabular*}
\end{table*}
\subsection{Dataset}
\label{IV-B}
We evaluate our method on the widely used 3D-RES dataset,  ScanRefer \cite{ref21}, which comprises 51,583 natural language expressions referring to 11,046 objects in 800 ScanNet \cite{ref16} scenes. The \textbf{Unique} subset of ScanRefer contains samples where only one unique object from a certain category matches the description. The \textbf{Multiple} subset contains ambiguous cases where there are multiple objects of the same category.

The evaluation metric is the mean IoU (mIoU) and Acc@$k$IoU, which represents the fraction of descriptions whose predicted mask overlaps with the ground truth with IoU $> k$, where $k \in \{0.25, 0.5\}$.

\subsection{Quantitative Comparison}
\label{IV-C}
Results of 3DResT and baselines are shown in Table~\ref{tab1}, where we first compare 3DResT with the fully supervised baseline and RefTeacher. The first observation is that 3DResT can achieve obvious improvement over the baselines. 3DResT outperforms the Supervised method by +18.00\% in the 1\% labeled data \textbf{Unique} part, +6.01\% in the 1\% labeled data \textbf{Multiple} part and +8.34\% in the 1\% labeled data \textbf{Overall} part. Besides, 3DResT outperforms RefTeacher in different data portions. Specifically, in the 1\% labeled data \textbf{Unique} part, 3DResT outperforms RefTeacher by +5.37\%. The second observation is that as the number of labeled data decreases, the improvement of 3DResT increases, demonstrating its effectiveness and suitability for scenarios with limited data. Specifically, the improvements are +2.80\%, +3.33\%, +8.37\%, and +8.34\% for 10\%, 5\%, 2\%, and 1\% labeled data, respectively. 

Our 3DResT can help the 3D-RES model effectively exploit unlabeled data to obtain competitive performance against the baseline methods.

\subsection{Ablation Study}
\label{IV-D}
\begin{table}
\centering
\caption{Ablation study of Quality-Driven Dynamic Weighting and Teacher-Student Consistency-Based Sampling on 1\% labeled data.}
\label{tab2}
\begin{tabular}{ccccc}
\toprule
\multirow{2}{*}{\textbf{\small QDW}} & \multirow{2}{*}{\textbf{\small TSCS}} & \textbf{\small Unique} & \textbf{\small Multiple} & \textbf{\small Overall} \\ 
\cmidrule(lr){3-5}
                                     &                                     & \textbf{\small mIoU}   & \textbf{\small mIoU}     & \textbf{\small mIoU}     \\ 
\midrule
                                     &                                     & {\small 51.03}          & {\small 17.02}          & {\small 23.62} \\
\checkmark                           &                                     & {\small 50.94}{\tiny -0.09}   & {\small 17.23}{\tiny +0.21}   & {\small 23.77}{\tiny +0.15} \\
                                     & \checkmark                          & {\small 55.03}{\tiny +4.00}   & {\small 18.06}{\tiny +1.04}   & {\small 25.23}{\tiny +1.61} \\
\rowcolor[HTML]{EFEFEF} 
\checkmark                           & \checkmark                          & \cellcolor[HTML]{FFD966}{\small 55.42}{\tiny +4.39} & \cellcolor[HTML]{FFD966}{\small 18.18}{\tiny +1.16} & \cellcolor[HTML]{FFD966}{\small 25.41}{\tiny +1.79} \\ 
\bottomrule
\end{tabular}
\end{table}
\textbf{The impact of TSCS and QDW.} In Table~\ref{tab2}, we provide a detailed ablation study to evaluate the impact of Teacher-Student Consistency-Based Sampling (TSCS) and Quality-Driven Dynamic Weighting (QDW) of 3DResT. These results clearly show that both TSCS and QDW contribute to improving performance, highlighting their value in this framework. TSCS outperforms QDW, achieving a notable improvement of +1.61\% with 1\% labeled data, demonstrating its ability to effectively utilize high-quality pseudo-labels. Besides, the results reveal that combining QDW with TSCS leads to even greater performance improvements. Specifically, the integration of these two designs achieves from 23.77 to 25.41 on 1\% labeled data. These results strongly validate the effectiveness of both TSCS and QDW.

\textbf{Comparisons of QDW with other designs.} Table~\ref{tab3} presents a comparison of QDW with other solutions for mitigating the issue of wasting useful information from low-quality pseudo-labels. In SSL, confidence filtering is a common approach to filter unreliable pseudo-labels. However, in referring tasks, where a pseudo-label contains only a single mask corresponding to the referred object, confidence filtering is a suboptimal solution. RefTeacher, a semi-supervised framework for 2D referring tasks,  employs a `predicted-weight' method that uses the predicted scores of pseudo-labels to weight the corresponding unsupervised loss. However, unlike detection tasks, which produce multiple predicted scores, such as confidence and classification scores, 3D-RES is a segmentation task and does not inherently generate such scores. In contrast, QDW effectively utilizes all low-quality pseudo-labels by dynamically adjusting their weights without incorporating any additional predictions. For instance, it achieves a performance improvement of +0.80\% on 10\% labeled data compared to the `predicted-weight' method. This demonstrates the ability of QDW to overcome the limitations of alternative methods in handling low-quality pseudo-labels.
\begin{table}
    \centering
    \caption{Comparisons of different methods for pseudo-label filtering on 10\% labeled data}
    \label{tab3}
    \setlength{\tabcolsep}{3pt}
    \begin{tabular}{lccc}
        \toprule
        \multirow{2}{*}{\textbf{\small Settings}} & \textbf{\small Unique} & \textbf{\small Multiple} & \textbf{\small Overall} \\
        \cmidrule(lr){2-4}
        & \textbf{\small mIoU} & \textbf{\small mIoU} & \textbf{\small mIoU} \\
        \midrule
        {\small Baseline} & {\small 68.48} & {\small 24.57} & {\small 33.09} \\
        {\small predicted-weight} & {\small 67.51}{\tiny -0.97} & {\small 24.00}{\tiny -0.57} & {\small 32.44}{\tiny -0.65} \\
        \rowcolor[HTML]{EFEFEF}
        {\small QDW} & \cellcolor[HTML]{FFD966}{\small 69.22}{\tiny +0.74} & \cellcolor[HTML]{FFD966}{\small 24.58}{\tiny +0.01} & \cellcolor[HTML]{FFD966}{\small 33.24}{\tiny +0.15} \\
        \bottomrule
    \end{tabular}
\end{table}
\begin{figure*}[!t]
    \centering
    \subfloat[\scriptsize Predictions of 3DResT with and without TSCS.]{
        \includegraphics[width=7.1in]{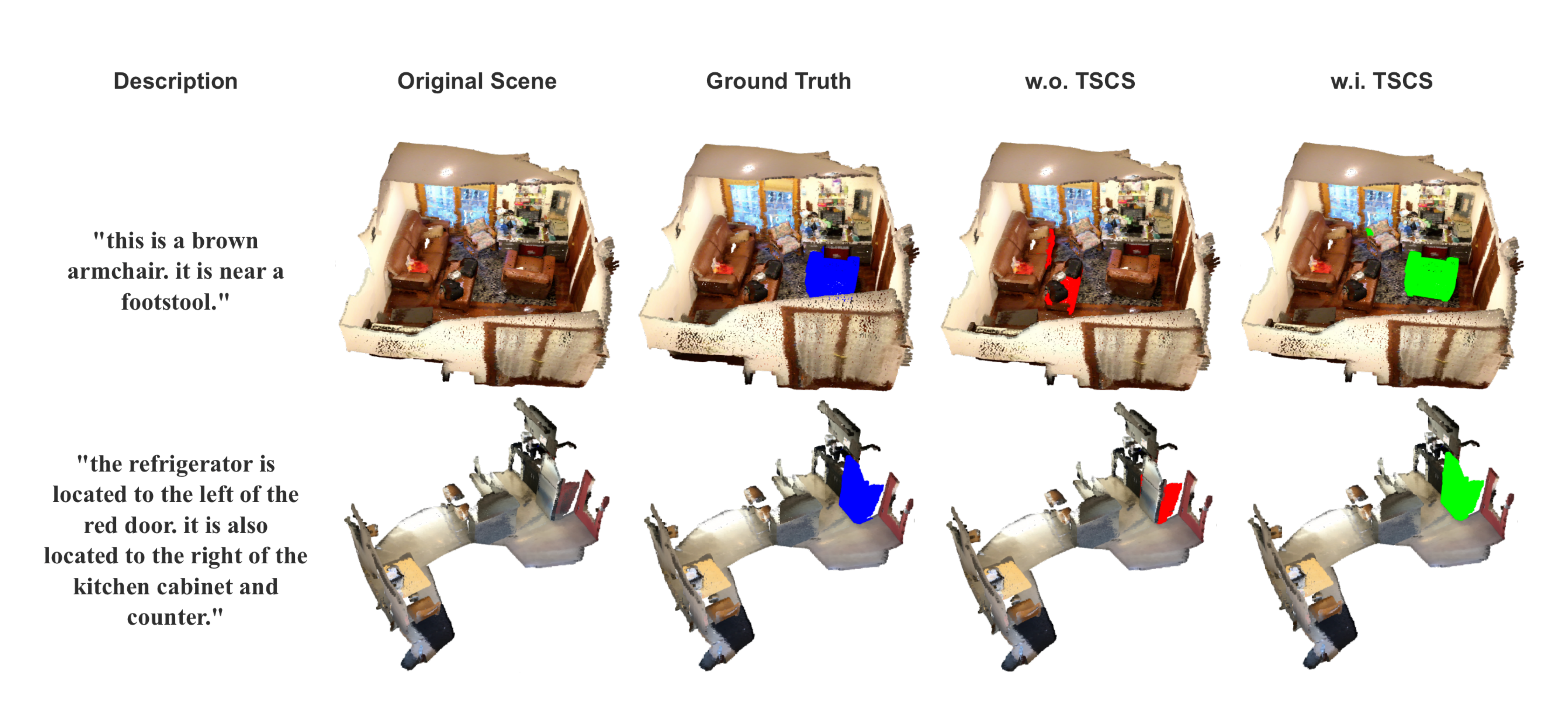}
        \label{fig:subfig_a}}
    \hfil
    \subfloat[\scriptsize Predictions of 3DResT with and without QDW.]{
        \includegraphics[width=7.1in]{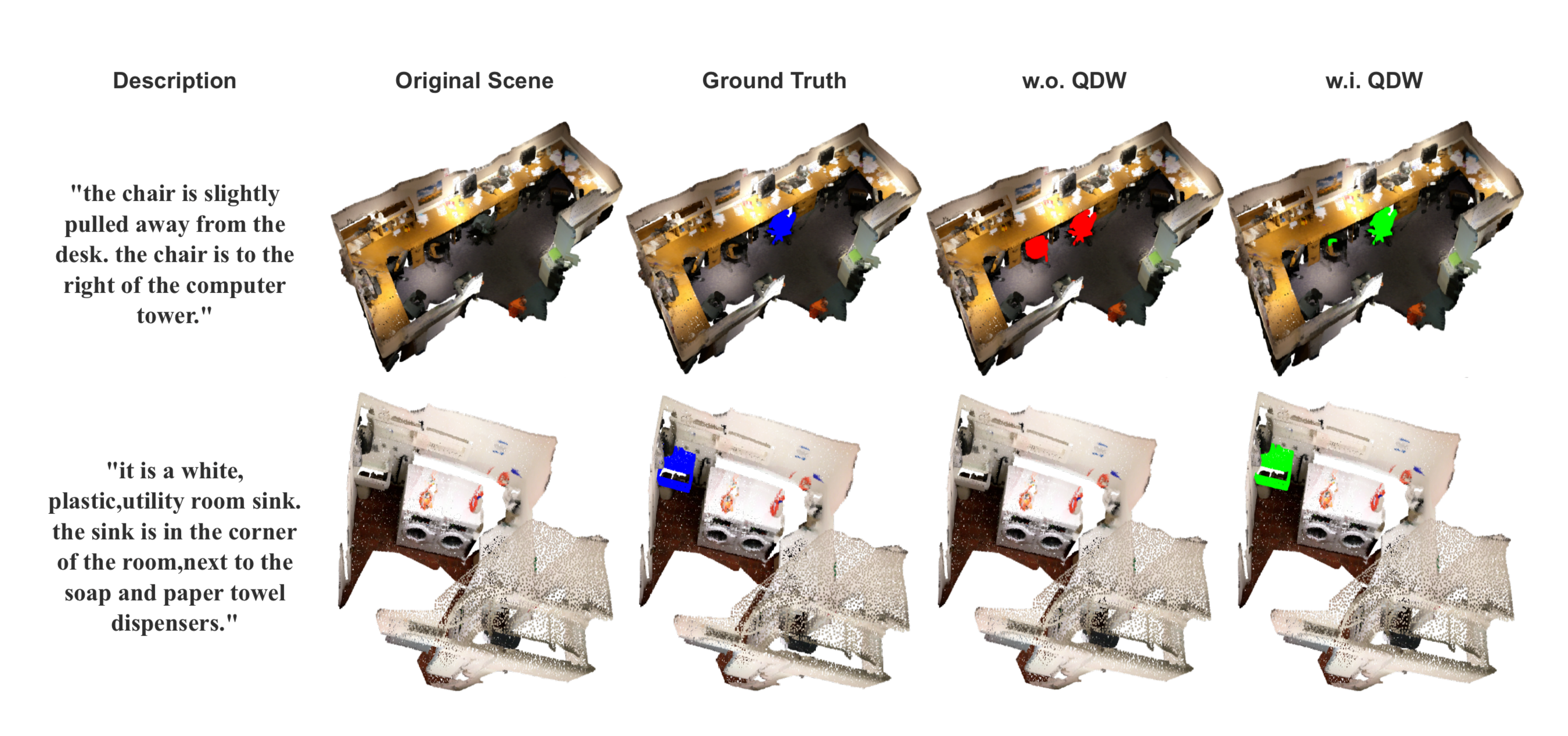}
        \label{fig:subfig_b}}
    
    \caption{Visualizations of 3DResT and fully supervised baselines. Subfigure (a) and (b) indicate that TSCS and QDW of 3DResT can obviously improve the quality of predictions under semi-supervised settings.}
    \label{fig4}
\end{figure*}

\textbf{Comparisons of diffenet periods in TSCS.} Investigating the optimal period for selecting high-quality pseudo-labels is crucial, as their quality varies across different periods. As shown in Table~\ref{tab4}, the model’s performance improves as the period progresses, peaking at `Mid' and then declining. Notably, our TSCS strategy achieves significantly better results, with a 1.61\% improvement compared to not selecting pseudo-labels at all.
\begin{table}
    \centering
    \caption{Comparisons of different periods of TSCS on 1\% labeled data. Total epoch is set to 30. Early, Mid and Late mean the early, intermediate and final stage of teacher-student mutual learning, respectively}
    \label{tab4}
    \setlength{\tabcolsep}{3pt}
    \begin{tabular}{lccc}
        \toprule
        \multirow{2}{*}{\textbf{\small Settings}} & \textbf{\small Unique} & \textbf{\small Multiple} & \textbf{\small Overall} \\ 
        \cmidrule(lr){2-4}
        & \textbf{\small mIoU} & \textbf{\small mIoU} & \textbf{\small mIoU} \\ 
        \midrule
        {\small w/o select} & {\small 51.03} & {\small 17.02} & {\small 23.62} \\
        {\small w select from Early} & {\small 52.06}{\tiny +1.03} & {\small 17.22}{\tiny +0.20} & {\small 23.98}{\tiny +0.36} \\
        \rowcolor[HTML]{EFEFEF}
        {\small w select from Mid} & 
        \cellcolor[HTML]{FFD966}{\small 55.03}{\tiny +4.00} & \cellcolor[HTML]{FFD966}{\small 18.06}{\tiny +1.04} & \cellcolor[HTML]{FFD966}{\small 25.23}{\tiny +1.61} 
        \\
        {\small w select from Late} & {\small 53.02}{\tiny +1.99} & {\small 17.58}{\tiny +0.56} & {\small 24.46}{\tiny +0.84} \\
        \bottomrule
    \end{tabular}
\end{table}

\subsection{Qualitative Comparison}
\label{IV-E}
\begin{figure*}[!t]
    \centering
    \includegraphics[width=7.1in]{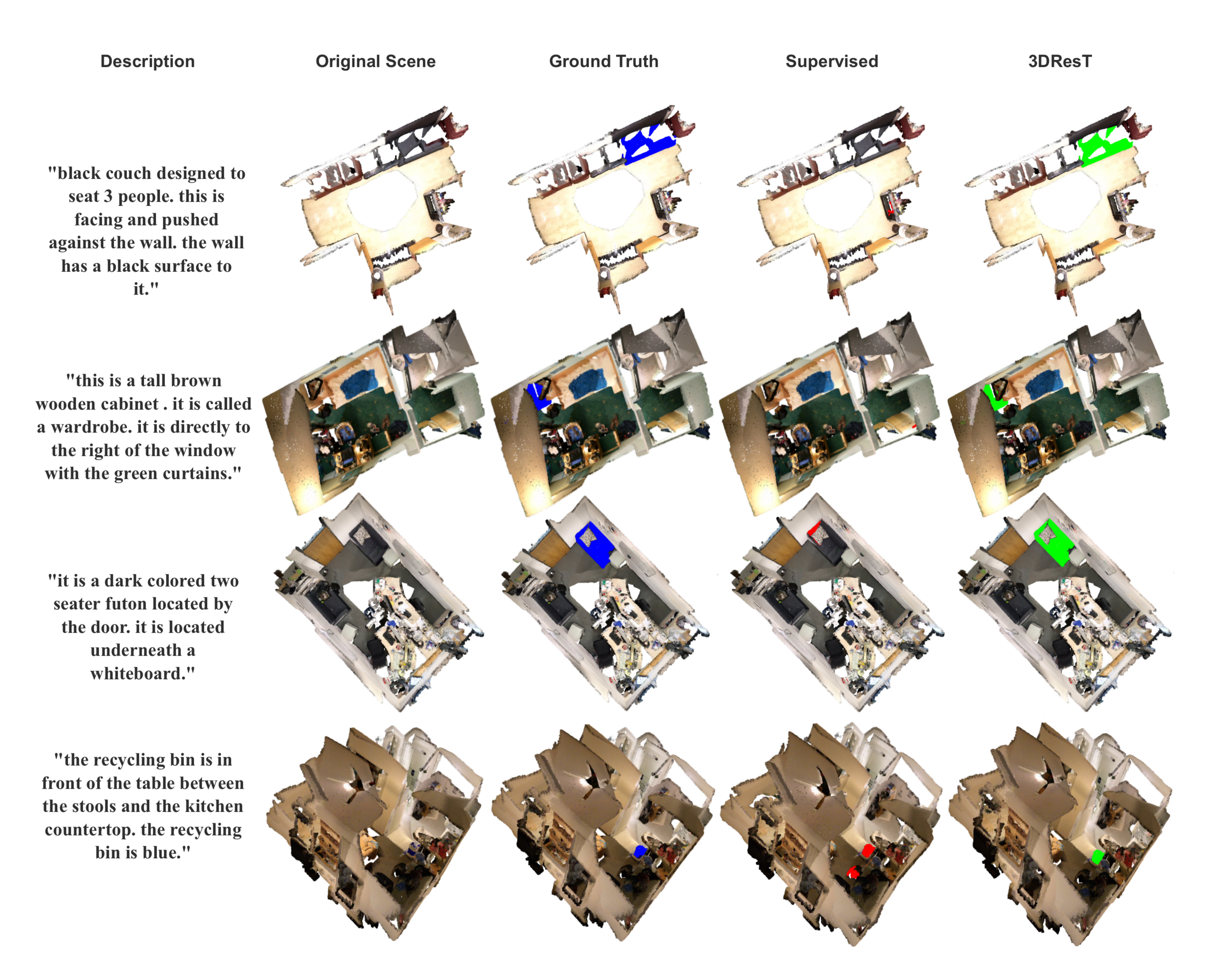}
    \caption{Comparisons of Supervised method and 3DResT. The blue mask means the ground truth, the red mask means the prediction of supervised method and the green mask means the prediction of 3DResT method}
    \label{fig3}
\end{figure*}
In this subsection, we present qualitative comparisons on the ScanRefer validation set, showcasing the remarkable discriminative ability of our 3DResT model. As shown in Fig. \ref{fig3}, the predictions of 3DResT outperform those of the supervised method. Specifically, in the top two rows, 3DResT accurately segments the referred objects, the black couch and wooden cabinet, whereas the supervised method fails to recognize them. In the bottom two rows, 3DResT can fully segment out the referred seater futon and precisely recognize the recycling bin, whereas the supervised method cannot achieve this. In Fig. \ref{fig:subfig_a}, we further compare the predictions with and without TSCS. In the upper example, TSCS enables the precise recognition of the referred armchair, which remains unrecognized without it. In the lower example, TSCS aids in segmenting the entire refrigerator, which cannot be fully recognized without it. Besides, Fig. \ref{fig:subfig_b} demonstrates the effectiveness of QDW. These visualizations clearly show that both our strategies significantly enhance the quality of the predictions.

\section{Conclusion}
In this paper, we present 3DResT, the first semi-supervised framework for 3D referring expression segmentation (3D-RES). Inspired by the success of semi-supervised learning in computer vision, 3DResT adopts the widely used teacher-student paradigm. This framework enables the effective utilization of large amounts of unlabeled data for training, substantially reducing annotation costs. Specifically, 3DResT addresses two key challenges in semi-supervised 3D-RES: inefficient utilization of high-quality pseudo-labels and wastage of useful information from low-quality pseudo-labels. To overcome these challenges, we propose two novel strategies: Teacher-Student Consistency-Based Sampling (TSCS) and Quality-Driven Dynamic Weighting (QDW). Extensive experiments demonstrate the superior performance of 3DResT when compared to both fully and semi-supervised baselines.

\end{document}